\documentclass{article}

\usepackage{arxiv}

\usepackage[utf8]{inputenc} 
\usepackage[T1]{fontenc}    
\usepackage{hyperref}       
\usepackage{url}            
\usepackage{booktabs}       
\usepackage{amsfonts}       
\usepackage{nicefrac}       
\usepackage{microtype}      
\usepackage{lipsum}		
\usepackage{graphicx}
\usepackage{doi}
\usepackage{dirtytalk}
\usepackage{setspace}
\usepackage{array}

\usepackage[%
  backend=bibtex      
 ,style=numeric-comp  
 ,sorting=none        
 ,sortcites=true      
 ,block=none
 ,indexing=false
 ,citereset=none
 ,isbn=true
 ,url=true
 ,doi=true            
 ,natbib=true         
]{biblatex}
\title{Reversing the Twenty Questions game}

\date{November 30, 2021}	

\author{{Parth Parikh} \\
	Department of Computer Science\\
	North Carolina State University\\
	Raleigh, NC 27606 \\
	\texttt{pmparikh@ncsu.edu} \\
	\And
	{Anisha Gupta} \\
	Department of Computer Science\\
	North Carolina State University\\
	Raleigh, NC 27606 \\
	\texttt{agupta44@ncsu.edu} \\
}

\renewcommand{\headeright}{CSC 791}
\renewcommand{\undertitle}{CSC 791 - Natural Language Processing}

\hypersetup{
pdftitle={Reversing the twenty questions game},
pdfauthor={Parth Parikh, Anisha Gupta},
pdfkeywords={Twenty Questions Game, Query Reformulation, Passage Retrieval, Boolean Question Answering Model, Natural Language Inference},
}

\begin{document}
\maketitle

\begin{abstract}
Twenty questions is a widely popular verbal game. In recent years, many computerized versions of this game have been developed in which a user thinks of an entity and a computer attempts to guess this entity by asking a series of boolean-type (yes/no) questions. In this research, we aim to reverse this game by making the computer choose an entity at random. The human aims to guess this entity by quizzing the computer with natural language queries which the computer will then attempt to parse using a boolean question answering model. The game ends when the human is successfully able to guess the entity of the computer's choice.
\end{abstract}

\keywords{Twenty questions game \and Query Reformulation \and Passage Retrieval \and Boolean Question-Answering Model \and Natural Language Inference}

\section{Introduction}
For our course project, we aim to reverse the roles of the computer and human, such that \textit{the computer will act as an answerer and a human as a questioner}. In the past, no such study has been conducted as this problem presented sophisticated challenges of Natural Language Inference and Textual Entailment. However, with the advent of transformer-based machine learning techniques such as BERT \cite{devlin2018bert}, RoBERTa \cite{liu2019roberta}, GPT-2 \cite{radford2019language}, and datasets such as BoolQ \cite{clark2019boolq}, such a model can be constructed.

As this problem has not been formally defined, our goal is to formalize it and present preliminary results regarding the same. Furthermore, while there are several pre-trained question-answering models that select the start and end points of a corpus containing an answer, a simple yes/no answering task is surprisingly challenging and complex. A model for such a task would have to examine entailment as well as investigate if the corpus makes a positive answer to the question unlikely, even if it doesn't directly state a negative answer \cite{clark2019boolq}. Our reverse Akinator model could be used for any sort of factual checker to examine whether a statement is true or not, given a knowledge corpus.

\section{Methodology}

\subsection{History}
Historically, Twenty Questions has been a popular multi-player parlour game wherein some participants would act as the \textit{questioners} and the others would be the \textit{answerers}. The answerers would come up with a random entity which the questioners would then try and deduce by asking a series of yes/no questions. A $19^{th}$ century rule-book \cite{walsorth1882twenty} details the format of the game and introduces the concept of umpires (who resolve any dispute) and captains (an official spokesperson). Interestingly, though the rule book never constrained the \textit{subject} (guess), every Sunday, it was mandatory for the participants to pick an object, person, or thing mentioned in the Bible.

Constrained versions of the game soon became popular and a variant known as the \textit{animal, vegetable, minerals} was widely played in parlours. As constraints produced tractability, one of the earliest computerized implementations of this game solely used \textit{Animals} as its subject \cite{TheAnimalGame}. This game was part of the \textit{101 BASIC Computer Games} (1973). Around 1988, \textit{20Q} created by Robin Burgener emerged. This version used an artificial neural network to answer questions based on a human's interpretation of that question. Today, popular internet-based variants such as \textit{Akinator} deals with a wide category of entities and includes \textit{Probably}, \textit{Probably not} and \textit{Don't know} as potential answers for a human.


\subsection{Entity Formulation and Pronoun Resolution}

Our proposed model starts by selecting a random Wikipedia page of a named entity. This entity acts as our model's main entity - \textit{the guess}. These random Wikipedia pages can be extracted by passing SPARQL queries to Wikidata \cite{vrandevcic2014wikidata}. The model then accepts natural language queries from a user. As the first step, each of these queries undergoes a basic pronoun resolution wherein a pronoun gets replaced with the model's main entity. For example, the model is likely to predict better results if we formulate the query in the following manner -\\

\centerline{Is \textbf{it} an animated character? $\to$ Is \textbf{Mickey Mouse} an animated character?}

This step ensures that our model does not easily get confused when it sees another entity with a similar context.

\subsection{Paragraph/Sentence Retrieval}
\label{paragraph-sentence-retrieval}
To obtain a relevant passage from the entity's Wikipedia text, we require a passage retrieval phase. Here, \textit{relevance} can be defined as a passage from the main entity's text-body which unambiguously answers a boolean-type query. For example - \\

\centerline{Is \textit{Mickey Mouse} a comic book character?}

\begin{center}
\begin{minipage}{.7\linewidth}
\say{Beginning in 1930, \textbf{Mickey has also been featured extensively in comic strips and comic books}. The Mickey Mouse comic strip, drawn primarily by Floyd Gottfredson, ran for 45 years. Mickey has also appeared in comic books such as Mickey Mouse, Disney Italy's Topolino and MM – Mickey Mouse Mystery Magazine, and Wizards of Mickey.} \\- \textit{From the Wikipedia page of Mickey Mouse (paragraph 3)}
\end{minipage}
\end{center}

As mentioned in \cite{jurafsky2018speech}, a trivial solution to this problem would be to perform sentence segmentation on the entire Wikipedia page and pass all the sentences to the question answering model. However, this can significantly affect the computational complexity as certain phases in BERT such as the \textit{multi-headed attention layer} requires $n^2\cdot d + n\cdot d^2$ operations (here $n$ is the sequence length and $d$ is the depth) \cite{Kaiser}.

A sophisticated variant would be to rank the passages based on the query and retrieve the first $N$ passages. We can use a ranking function such as Okapi BM25 \cite{schutze2008introduction} for such a task. However, as \cite{schutze2008introduction} uses a bag-of-words-based approach, its rankings can be too literal and devoid of any implicit context. To resolve this, we introduce a hybrid approach wherein a large subset of $N_1 \subseteq P$ passages is retrieved using BM25 and a much smaller subset $N_2 \subseteq N_1$ is then obtained using \textit{Siamese BERT-Networks} \cite{reimers2019sentence}. Here, sentences/paragraphs are mapped to a dense vector representation using transformer networks such as BERT, which can then be compared using cosine similarity. We plan on embedding the query $Q$ and comparing it against the embeddings of each $n \in N_2$, keeping a track of the top $N$ passages. A Python library - \textit{Sentence Transformers} \cite{Reimers_Gurevych} provides pre-trained models for this task.

The above mentioned model uses a \textit{sparse-first search} mechanism wherein we retrieve the $N_1$ documents using a statistical approach which is followed by a neural model. The drawback of this is that we may propagate errors from the document retrieval phase. That is, if we retrieve the wrong documents then it might affect the performance of the Transformer models. To mitigate this, Facebook Research developed \textit{Dense Passage Retrieval} \cite{karpukhin2020dense} which uses the concept of indexing phrases using a dual-encoder framework. Here, they enumerate a document for all phrases in that document and use a phrase encoder to embed each phrase in vector space. The queries are mapped to the same vector space and Nearest Neighbour Search is used to obtain the most relevant answers. 



\subsection{Boolean Question Answering Model}
To guess the boolean-type response, we propose a transformer-based model which takes as its input a query and $N_2$ relevant paragraphs. We plan on experimenting with a BERT model pre-trained on entailment tasks and fine-tuned using the BoolQ dataset \cite{clark2019boolq}. \cite{clark2019boolq} showed that the highest accuracy is obtained when we pre-train models on entailment tasks that have large datasets (such as MultiNLI \cite{williams-etal-2018-broad} and SNLI \cite{bowman-etal-2015-large}) and fine-tuning them on BoolQ's dataset. 

While playing games with Akinator, we observed that a certain class of questions can be answered using knowledge repositories such as Wikidata and DBpedia \cite{auer2007dbpedia}. These questions involve highly distinguishing characteristics of the entity such as its gender, species, hypernyms, and significant others. 

\section{Experiments}
As mentioned in Report 1's evaluation section, we verified our model's performance by playing it against the pre-existing Akinator using the Python library \textit{akinator.py} \cite{akinator.py}. This library acts as the original Akinator, posing questions to our model and trying to guess which entity our model has in mind. The number of questions asked by the Akinator is not constrained in our experiments. We only stop the game once the Akinator guesses an entity with a probability greater than 80\%. 

\subsection{Akinator API}
The Akinator API \cite{akinator.py} allows us to access the Akinator’s top guesses at a particular time, with a guess probability and a rank. The first guess is used to evaluate if the Akinator won (that is, if the Akinator was able to guess the answer correctly). The API also allows us to go back to a previous question and change our answers. Furthermore, we are able to select a nature of the entities that we want to guess. This comprises of language options (such as English, Chinese, German), and entity types (like animals, characters, and objects).

\subsection{Baseline model}
Our initial baseline model answers the Akinator’s questions at random with \textit{Yes}, \textit{Probably}, \textit{I don’t know}, \textit{Probably not} and \textit{No}. However, when we performed our experiments, we observed that too many \textit{i don’t know} or \textit{probably yes/no} responses would make the Akinator guess something along the lines of \textit{a guy who plays randomly} (this statement is one of Akinator's named entity which it assigns to anyone who guesses randomly). So we allocated these responses a much lower probability of $0.05$ each, and distributed the remaining probabilities uniformly among the rest of the answer options, such that the baseline model could make a probabilistic random choice.

An entity only shows up when it is within the top few guesses of the Akinator. From our experiments on our initial baseline model, we hardly ever see the desired item show up in the list of top few guesses of the Akinator.

\begin{figure}
    \centering
    \includegraphics[scale=0.3]{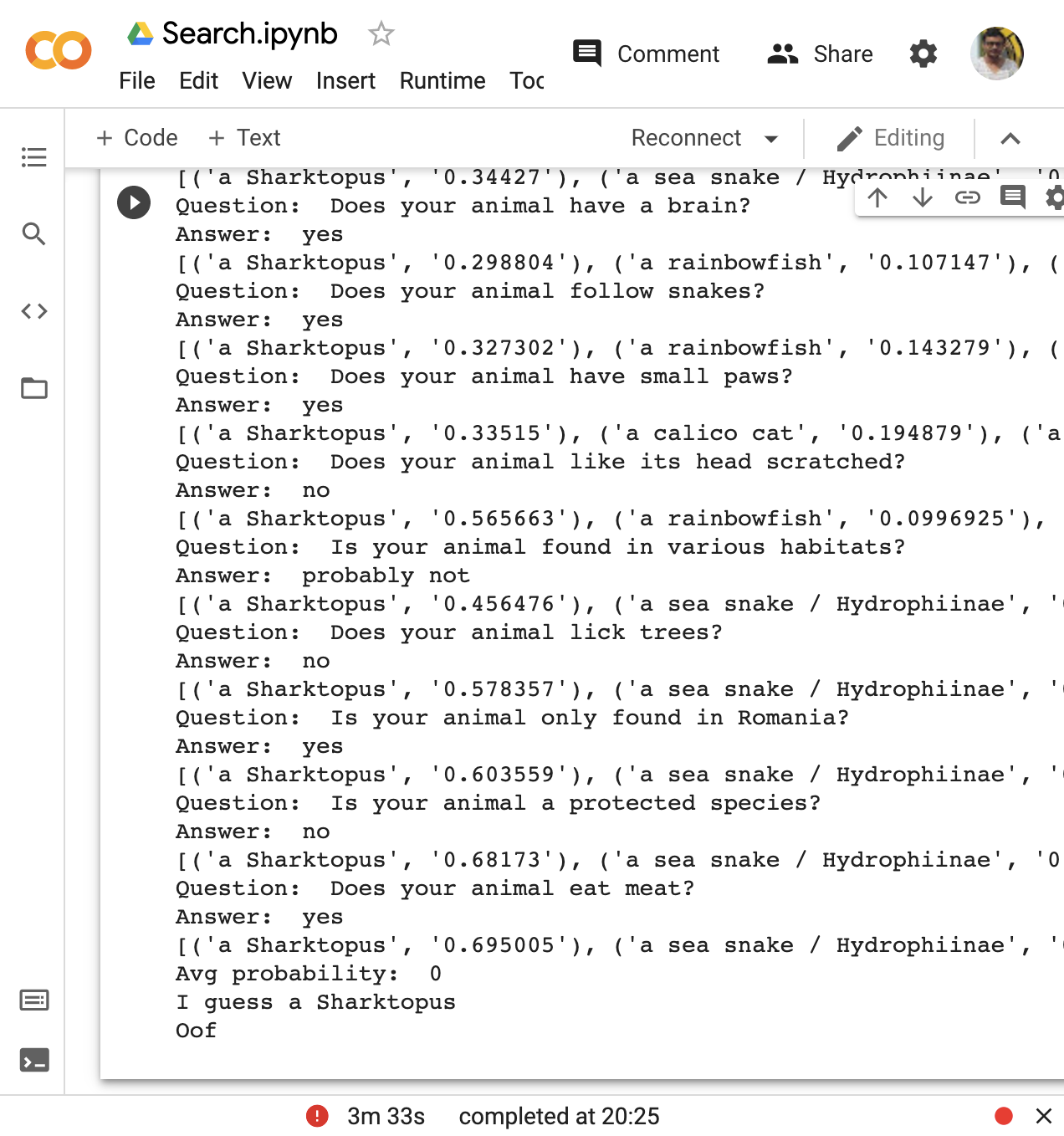}
    \caption{Random baseline model results}
    \label{fig:random-baseline}
\end{figure}

From the results shown in Figure \ref{fig:random-baseline}, we see that the Akinator’s guess converges to a \textit{Sharktopus} with a final probability $> 80\%$. However, the guess is incorrect, as is expected, since it’s a random model. The desired animal (\textit{Cheetah}) never features in the Akinator’s guess list. In this model, the correct answer can only show up in the list of top guesses by chance, and this happens very rarely.

We performed some preliminary analysis using anaphora resolution on the questions asked by the Akinator. However, in some cases (ex. Table\ref{table:coref-resolution-dilemma}), the extracted answer excerpts are more unrelated to the question after applying anaphora resolution to the question. As part of our preliminary analysis, we also explored the BERT Question Answering model. However, based on manual inspection of the results, the excerpts extracted using the BERT Question Answering model are less relevant to the question than that extracted using our pipeline. This could be supported by Reimers et al.'s work \cite{reimers2019sentence}, where they show that averaging the \texttt{[CLS]} tokens for the BERT embeddings “...yields rather bad sentence embeddings, often worse than averaging GloVe embeddings”.

\subsection{Improved Model}
For our improved model, we implemented the \textit{Okapi-BM25/SBERT} pipeline proposed in Section \ref{paragraph-sentence-retrieval}. We fixed $N_1$ to 100 and $N_2$ to 5. For our current experiments, our pipeline outputs these top $N_2$ \textit{most similar} excerpts that answers the Akinator’s question at each step, and lets the human developer answer a \textit{Yes}/\textit{No} based on these top five excerpts.

\begin{figure}
    \centering
    \includegraphics[scale=0.6]{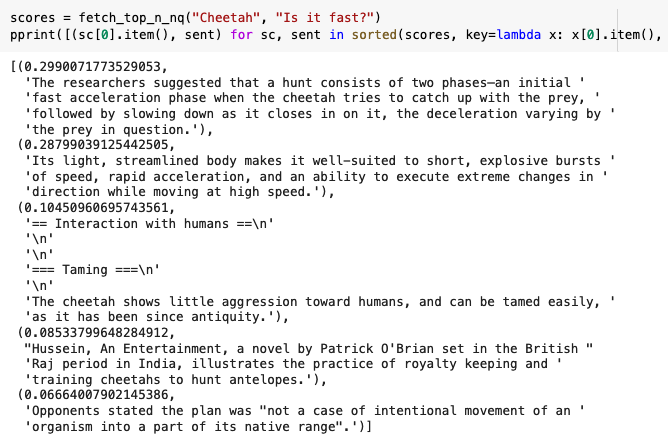}
    \caption{Sample results using improved pipeline}
    \label{fig:example-improved-pipeline}
\end{figure}

An example output of the same is shown in Figure \ref{fig:example-improved-pipeline}.

\subsubsection{Constraining the domain}
\begin{figure}
    \centering
    \includegraphics[scale=0.3]{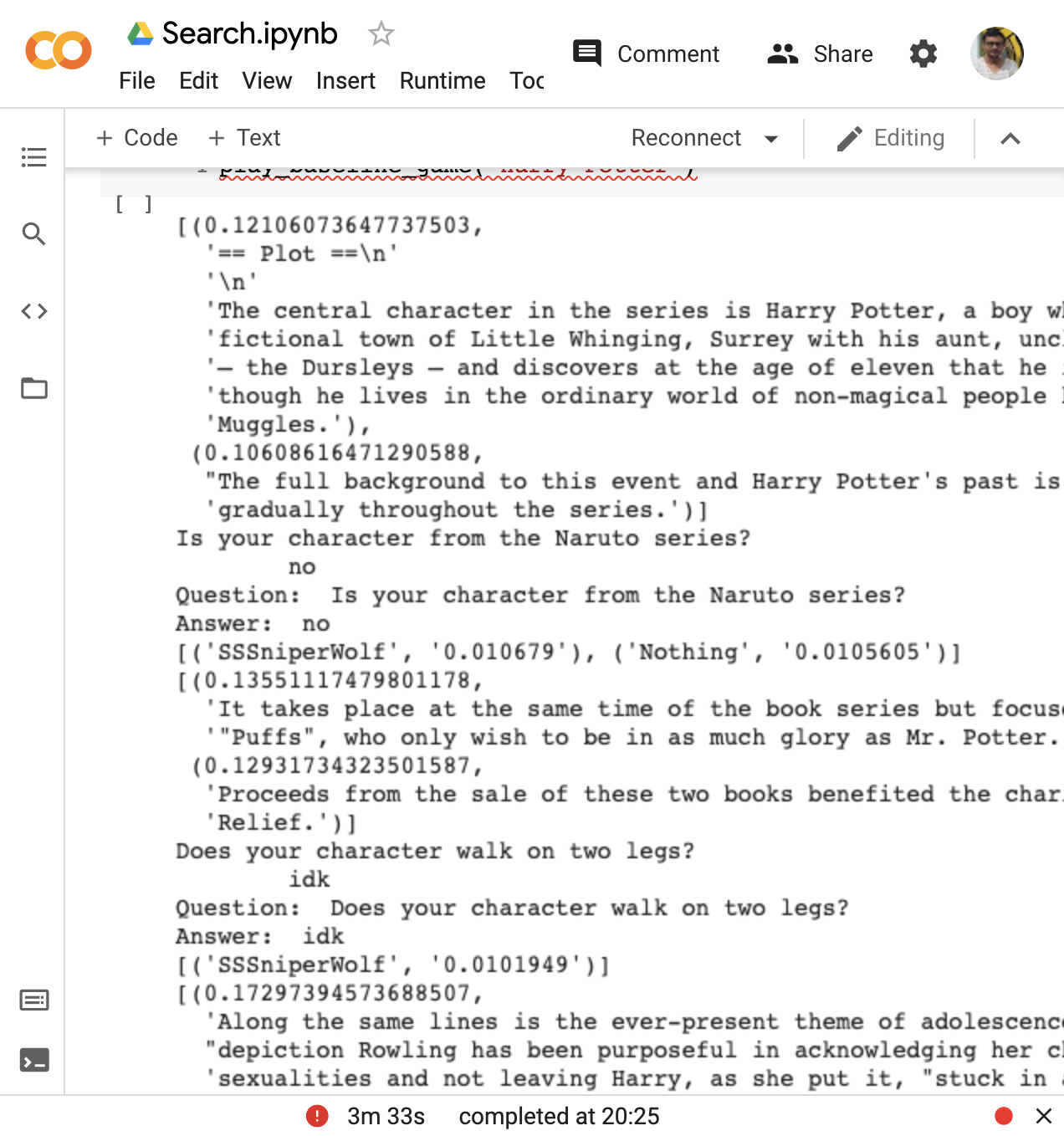}
    \caption{Answer excerpts extracted for fictional character Harry Potter}
    \label{fig:harry-potter-demo}
\end{figure}

Our initial experiments using the aforementioned pipeline did not produce good results for general entities, including movie characters such as \textit{Harry Potter}, as can be observed from the example in Figure\ref{fig:harry-potter-demo}. This is often because the complexity of information is more for such characters, with both real life and reel life data, as well as information about a lot of other characters/persons documented in the Wikipedia articles. Given the lack of access to knowledge graphs, trivia questions are more difficult for our model to answer. We thus constrain our domain to English animal names. The Wikipedia articles corresponding to a certain animal usually only talks about this animal and does not have a lot of content on other animals or information that requires a knowledge base for answering questions, thus making our problem more tractable for our purposes. 

\subsubsection{Simple Wikipedia pipeline}
In a lot of cases, our model was unable to distinguish between excerpts that referred to the actual animal and cultural references to that animal. For instance, when asked the question ‘\textit{Does your animal [cheetah] still exist?}’, the following text excerpt is extracted from our cheetah wikipedia corpus with a very high confidence score:

\begin{quote}
\textit{The Bill Thomas Cheetah American racing car, a Chevrolet-based coupe first designed and driven in 1963, was an attempt to challenge Carroll Shelby's Shelby Cobra in American sports car competition of the 1960s era. Because only two dozen or fewer chassis were built, with only a dozen complete cars, the Cheetah was never homologated for competition beyond prototype status; its production ended in 1966.
}    
\end{quote}

Based on this excerpt, the yes/no model answers ‘\textit{No}’, indicating that cheetahs are extinct, which immediately throws off the Akinator and it starts thinking of types of dinosaurs. However, we do not wish to completely disregard cultural references - these excerpts are helpful when questions such as ‘\textit{Is there a car named after your animal?}’ are posed by the Akinator. To avoid confusing our pipeline with such cultural references that do not directly relate to the animal in general, we ask the same question to the Simple Wikipedia corpus for our animal, and append the answer we get from here to the answer excerpt we get from the original Wikipedia article. If the average text confidence scores for the Simple Wikipedia and original Wikipedia articles is less than one standard deviation of the average negative sample scores on the same question, the pipeline outputs ‘idk’ as a response. Otherwise, we output yes/no based on our boolean answer model prediction on a combination of text answers from Simple Wikipedia and original Wikipedia.

\subsubsection{Detecting comparisons}
For certain questions such as ‘\textit{Is your animal smaller than a human?}’ or ‘\textit{Is your animal bigger than your hand?}’, the model requires real world knowledge to provide accurate answers - \textit{How tall is a regular human?} and \textit{How big is an average human hand?}. Handling such cases is challenging and beyond the scope of this project. However, to mitigate the consequences of answering these questions incorrectly, we inspect the question for ‘\textit{comparison}’ words included in NLTK’s \texttt{comparative\_sentences} dictionary, such as ‘\textit{smaller}’, ‘\textit{shorter}’, etc. If the question contains such comparison words, the pipeline outputs an `idk' response. If a correct answer to this question was expected to boost the probability of our animal in the Akinator’s guess list, it might reduce the probability by a bit, but not as dramatically as an incorrect answer would lower the probability.

\subsubsection{Converting answer excerpts to Yes/No}
\begin{figure}
    \centering
    \includegraphics[scale=0.3]{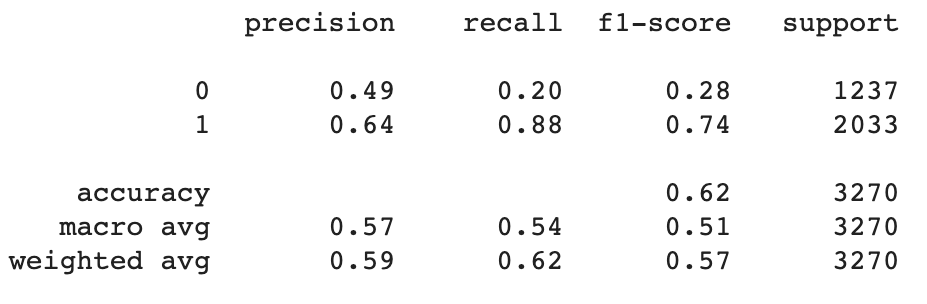}
    \caption{Baseline results obtained after converting answer excerpts to Yes/No labels for the BoolQ dataset}
    \label{fig:boolq-baseline-results}
\end{figure}

\paragraph{Multilayer Perceptron Classifier}
For Report 1, we designed a baseline model for this classification task. We trained a Multilayer Perceptron Classifier model on the BoolQ dataset \cite{clark2019boolq} to predict a Yes/No answer, given a question, and an answer excerpt from a passage. Each question and answer excerpt was first converted to an embedding vector by computing the GloVe embeddings of each token and averaging these over all the tokens. NLTK’s TweetTokenizer \cite{loper2002nltk} was used for word tokenization. The average of the question and excerpt embedding was then performed to obtain a semantic embedding representing the QnA phase, which was passed as an input to our classifier. The results of this model are shown in Figure \ref{fig:boolq-baseline-results}.

\begin{figure}
    \centering
    \includegraphics[scale=0.4]{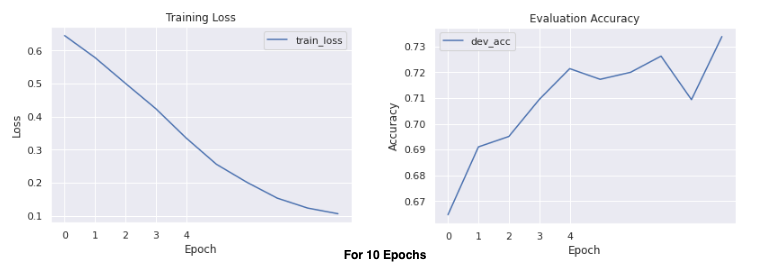}
    \caption{Training loss and Dev accuracy after fine-tuning on DistilBERT for 10 epochs}
    \label{fig:10-epochs}
\end{figure}

\begin{figure}
    \centering
    \includegraphics[scale=0.4]{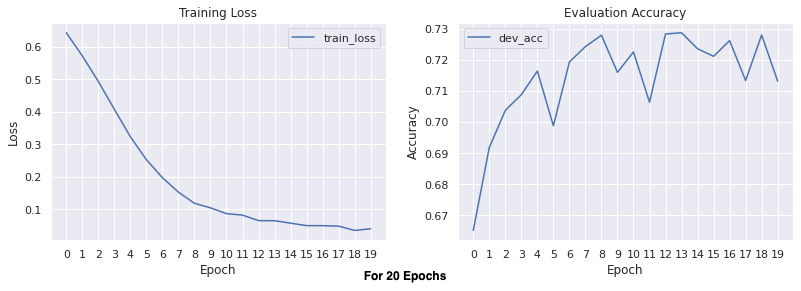}
    \caption{Training loss and Dev accuracy after fine-tuning on DistilBERT for 20 epochs}
    \label{fig:20-epochs}
\end{figure}

\paragraph{DistilBERT}
From our results we see that the model has a low F1 score for prediction of \textit{No}. For Report 2, we improved upon the Multilayer Perceptron Classifier model by architecturing an entailment model and fine-tuning it on the BoolQ dataset. The authors of the BoolQ paper observed their best performance by using the pretrained BERT-large transformer model and fine-tuning it on their dataset. For our model, we experimented with DistilBERT - a lighter version of BERT with 97\% of its language understanding capabilities and 60\% faster. To train this model, we utilized its SequenceClassification model with batch size of 32, learning rate of $10^{-5}$ and Adam optimization for stochastic gradient descent with gradient clipping. This model was fine-tuned on the BoolQ dataset. We trained it for 3 different epochs - 5 (35 minutes), 10 (110 minutes), 20 (230 minutes), and observed that 5 epochs severely overfitted on "Yes" response. However, 10 epochs reduced the overfitting, decreased the training loss to nearly 10\% and provided a dev accuracy of 73.3\%. Moreover, with 20 epochs, we experienced a  severe overfitting on BoolQ with the model having difficulty converging due to a high learning rate. The figures detailing the same are figures \ref{fig:10-epochs} and \ref{fig:20-epochs}.

\begin{figure}
    \centering
    \includegraphics[scale=0.4]{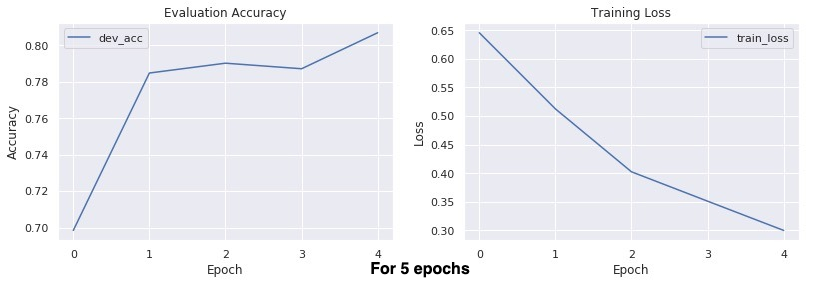}
    \caption{Training loss and Dev accuracy after fine-tuning on RoBeRTa for 5 epochs}
    \label{fig:roberta-5}
\end{figure}

\begin{figure}
    \centering
    \includegraphics[scale=0.4]{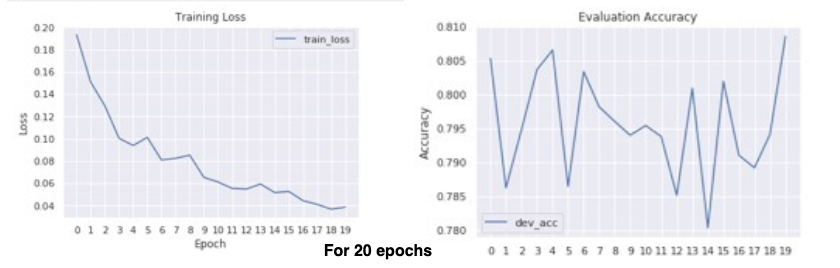}
    \caption{Training loss and Dev accuracy after fine-tuning on RoBeRTa for 20 epochs}
    \label{fig:roberta-20}
\end{figure}

\paragraph{RoBeRTa-base}
For our model, we also experimented with RoBeRTa-base transformer which is an improvement over the BERT-large transformer, as it performs dynamic masking with 500 thousand optimization on batch sizes of 8000 (for comparison, BERT has batch size of 256), and is pretrained on 160 GB of data. It removes the next-sentence prediction as seen in BERT, and trains each batch over longer sequences of data. We kept the learning rate and Adam Optimization same as our  DistilBERT implementation. After fine-tuning the RoBeRTa-base transformer on BoolQ for 20 epochs (with batch size of 32), we noticed a training loss of 4\% and development-set accuracy of 80.7\%. This is a significant improvement from the DistilBERT implementation which consisted of a development-set accuracy of 73.3\%. Figures \ref{fig:roberta-5} and \ref{fig:roberta-20} display the training loss and development-set accuracy for 5 epochs.

\subsubsection{Negative Sampling}
\begin{figure}
    \centering
    \includegraphics[scale=0.4]{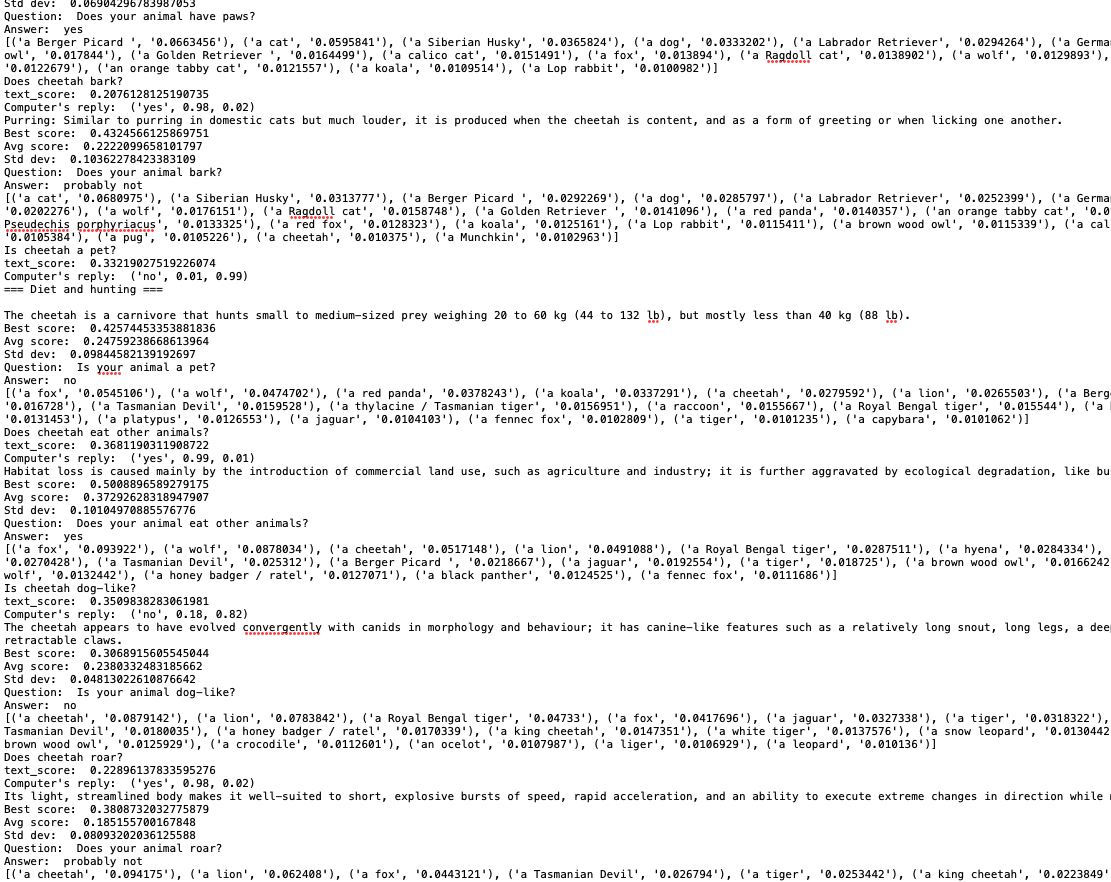}
    \caption{Example game for animal cheetah}
    \label{fig:cheetah-game}
\end{figure}

Usually, if an animal does not possess a certain characteristic, it is not mentioned in the Wikipedia article for that animal. In such cases, the results obtained by BM25 and cosine similarity might be misleading. Despite computing similarity scores for the relevant answers, the threshold determining which answer is appropriate for the question could be difficult to determine. For instance, if the entity is a \textit{cheetah} and we want to find out if \textit{it is an animal that can be used in shows}, the most relevant answer from the Wikipedia article for \textit{cheetah} is \textit{The cheetah has been widely portrayed in a variety of artistic works.} However, this does not answer the original question in the sense in which it was asked. To tackle this challenge, if we do not get \textit{Yes} as an answer to our question on the correct animal, we propose a negative sampling technique where we design a taxonomy of animals and select one entity at random from each broad category and treat these as negative samples to our model. The taxonomy uses a sample of well-known animals from ten broad categories - amphibians, birds, carnivores, domestic, fish, herbivores, invertibrates, mammals, primates and reptiles. We ask the same question with respect to all these negative samples and select the topmost ranking answer excerpts for each animal. We then compare the scores of these top answers with our current animal. 
\\ \\
If the score of a negative sample is more than one standard deviation of that of our top answer (for the correct animal), it reflects low probability of finding the answer in the Wikipedia file for our correct animal. In this case, we check if the score of our top answer is within one standard deviation of the mean score of all the negative samples considered - if not, it will indicate that the score for our top answer is really low and there is no mention of the answer in the Wikipedia file, which would mean that the model doesn't know the answer and should output \textit{idk}. Otherwise, if the top answer score is not within one standard deviation of the best negative sampling score but more than the mean score, we output \textit{probably yes} if the BoolQ yes/no answering model outputs \textit{yes} or \textit{probably no} if the yes/no answering model outputs \textit{no}. An example of how this works is shown in Table \ref{table:negative-sampling-table}. An example game excerpt incorporating negative sampling with the BoolQ outputs is shown in Figure \ref{fig:cheetah-game}.

\begin{table}
\begin{center}
\begin{tabular}{|m{5em}|m{25em}|m{5em}|m{5em}|}
\hline
Entity name & Sentence                                                                                                                                                                                                                                                                                                                             & Probability & Positive Sample? \\ \hline
Cheetah     & They have been widely depicted in art, literature, advertising, and animation.                                                                                                                                                                                                                                                       & 0.17        & Yes              \\ \hline
Cheetah     & An open area with some cover, such as diffused bushes, is probably ideal for the cheetah because it needs to stalk and pursue its prey over a distance.                                                                                                                                                                              & 0.10        & Yes              \\ \hline
Dog         & In conformation shows, also referred to as breed shows, a judge familiar with the specific dog breed evaluates individual purebred dogs for conformity with their established breed type as described in the breed standard.                                                                                                         & 0.26        & No               \\ \hline
Dog         & In 2015, a study found that pet owners were significantly more likely to get to know people in their neighborhood than non-pet owners.Using dogs and other animals as a part of therapy dates back to the late 18th century, when animals were introduced into mental institutions to help socialize patients with mental disorders. & 0.17        & No               \\ \hline
Frog        & It is typically used when the frog has been grabbed by a predator and may serve to distract or disorient the attacker so that it releases the frog.                                                                                                                                                                                  & 0.19        & No               \\ \hline
Frog        & Frogs are used for dissections in high school and university anatomy classes, often first being injected with coloured substances to enhance contrasts among the biological systems.                                                                                                                                                 & 0.15        & No               \\ \hline
Penguin     & Several species are found in the temperate zone, and one species, the Galápagos penguin, lives near the Equator.                                                                                                                                                                                                                     & 0.11        & No               \\ \hline
Penguin     & In the 60s Batman TV series, as played by Burgess Meredith, he was one of the most popular characters, and in Tim Burton's reimagining of the character in the 1992 film Batman Returns, he employed an actual army of penguins (mostly African penguins and king penguins).                                                         & 0.09        & No               \\ \hline
Snail       & Snails have considerable human relevance, including as food items, as pests, and as vectors of disease, and their shells are used as decorative objects and are incorporated into jewelry.                                                                                                                                           & 0.15        & No               \\ \hline
Snail       & Land snails are known as an agricultural and garden pest but some species are an edible delicacy and occasionally household pets.                                                                                                                                                                                                    & 0.11        & No               \\ \hline
\end{tabular}
\end{center}
\caption{Negative Sampling}
\label{table:negative-sampling-table}
\end{table}


\subsubsection{Training improved Yes/No model using negative samples}
To further leverage answers extracted from the randomly selected negative samples, we hand-annotated 250 questions asked by the animal, for a list of 15 animals. We recorded the yes/no answers generated by our automated question answering pipeline, as well as the text score statistics (average, best and standard deviation) of the negative samples on the same question. We tried to train a model that aims to identify situations where the initial yes/no answer must be modified if the negative sample scores hint that the answer may not be present in the corpus for our animal. Given the limited number of hand-annotated samples, we used simple models like MLP, SVC and decision trees. However, most of the yes/no answers (>78\%) matched with the human annotations and did not need any correction, resulting in the model overfitting on the initial yes/no answer and not utilizing the negative sample score statistics to make an improved prediction. We believe that an increased number of hand-annotated samples will improve the predictive performance of such models and can be incorporated as an improvement step after obtaining the initial yes/no answer from the model.

\subsubsection{Detecting and fixing a detour}
Based on our experiments, we noticed that the Akinator’s guess list is extremely volatile and sensitive to all answers. Even if the correct animal shows up in the guess list with the highest probability, the answer to the immediate next question can reduce its probability drastically, to the point of it getting eliminated entirely from the guess list. Fortunately, the Akinator has a weak long term memory, giving more importance to recent answers. This helps the Akinator return to animals similar to the correct animal after taking a long detour, and the answer often converges to the correct animal after the Akinator recovers from the detour. However, this might take a long time, and we might hit the maximum number of questions (80) after which the Akinator throws an error.
We want to detect such detours early without allowing our pipeline to peek into the Akinator’s guess list at any given time. This is challenging, given that we do not know where the Akinator’s guesses are headed at any given time, and we are not aware of whether our past answers are correct or incorrect. We propose a technique to detect misleading answers using negative sampling results, and bring the Akinator back using positive samples - animals that are most similar to the correct animal. 

\paragraph{Negative sampling to detect a detour}
To judge which animals are similar/dissimilar to the correct animal, we extract the word embeddings for each animal in the negative sampling list and our vocabulary of animals, and compare these with the word embedding for the correct animal. We expect the embeddings for animals such as ‘dog’ and ‘cat’ to be more similar to each other and different from ‘crocodile’ and ‘giraffe’. We consider a fixed negative sampling list (sampled randomly) for the entire game. For each question that the Akinator asks, we answer yes/no for all the animals in our negative sampling list, as well the correct animal. We store W most recent yes/no answers for all animals in the negative sampling list and for the correct animal. After answering each question, we check to see if our last W yes/no answers have been too similar to an animal in the negative sampling list that is very dissimilar to the correct animal. If so, we report a detour.

\paragraph{Fixing a detour}
If we detect a detour, we inspect our animal vocabulary to identify N animals that are most similar to the correct animal. We call this our positive sampling list. We answer the next question with a majority yes/no vote from these positive samples. We do not answer every question this way because the Akinator is not likely to converge to the correct animal if we answer specific questions such as ‘\textit{Does your animal have spots?}’ incorrectly. Once we have fixed a detour, we empty the past W yes/no answers list for all animals in the negative samples - we do not want to apply this technique too early.

\section{Model Evaluation}
We use accuracy, recall, precision and F1 score on the BoolQ test set as the evaluation metric for the submodel used to convert extracted answers to \textit{Yes}/\textit{No}. We can evaluate the submodels on pre-existing benchmarks. GLUE \cite{wang2018glue} contains several tasks such as similarity, paraphrasing and inference tasks, and can be used to evaluate the quality of sentence embeddings used in our model. SuperGLUE \cite{wang2019superglue} can be used to test our question answering model. QNLI \cite{wang2018glue} dataset can be used to determine whether our selected answer excerpt contains the answer to the question posed by the Akinator. WNLI \cite{wang2018glue} can be used to evaluate our model's anaphora resolution performance, if we include this as a component of our final model. 

We hand-annotated answers to 250 questions and compared these answers to the yes/no outputs of our pipeline. The answers matched with an accuracy of 78.69\% and F1 scores 81.67\% (class no) and 74.61\% (class yes). Since it requires a lot of manual effort to hand-annotate these answers and play long games with the Akinator, we devised an approximate answering technique that guesses the correct yes/no answer for each question. This technique can only be applied to answers that result in the correct animal appearing in the Akinator’s guess list. If the probability of the correct animal in the Akinator’s guess list increases after answering a question, we estimate that answer to be a correct answer (correct answer equals the pipeline’s output). Otherwise, we mark the answer as incorrect (correct answer is the opposite of the pipeline’s output). Using this estimated correct answer, we labeled another 264 question-answer pairs that were automatically generated in games with the Akinator. 62.88\% questions were answered correctly, considering the expected correct answer as the ground truth. However, there might be inconsistencies in this ground truth. For instance, there have been instances of cheetahs being tamed in human history, and a cheetah is technically not able to roar - but the Akinator reduces the probability of ‘cheetah’ when it asks these questions and the pipeline answers correctly based on the Wikipedia article. So a probability reduction in the guess list may not always be indicative of an incorrect answer. The Akinator, at the end of the game, asks for the actual answer if it fails to identify the entity that the user had in mind, suggesting that it updates its knowledge by some sort of crowdsourcing, which may result in these anomalous results.

For further evaluation, we designed a couple of metrics - number of questions it takes the akinator to reconsider the correct animal after being thrown off by an incorrect answer (\textit{detour recovery time}) and the best probability of the animal in the Akinator’s guess list over the entire game (\textit{best guess probability}).

\paragraph{Detour recovery time}
We measure the time (measured by the number of questions) taken by the Akinator to recover from an incorrect answer that knocks off the correct animal from the guess list to the point where it is reintroduced in the Akinator’s guess list. Evaluating on our automated game results, we observe an average span of approximately 8 questions before the Akinator is able to come back on track. This gives us an intuition of how fast the model is able to redirect the Akinator’s focus - the lesser the detour recovery time, the better. A longer detour recovery time would indicate that the pipeline has answered incorrectly multiple times in succession, which might cause the Akinator to drift further away from the actual answer. The Akinator is able to come back on track eventually most of the time because it does not seem to have a strong long term memory and focuses more on recent answers. 

\paragraph{Best guess probability}
We record the highest probability with which the correct animal features in the Akinator’s guess list over the course of a single game. The average best guess probability for an experiment on 15 animals was 25.91\%. This is a relatively high probability, given that most of the times when the Akinator considers an animal in the guess list, it starts off with a probability of less than 1\%. 

We propose an additional metric for future implementation to get a better understanding of our pipeline’s performance - convergence rate. This metric could consider the initial probability (from the time that the correct animal shows up in the Akinator’s guess list) and the final probability (highest probability achieved by the Akinator for the correct animal), and the rate of this increase over the number of questions asked between the initial and final probability timestamps. If the correct animal disappears from the Akinator’s guess list, the convergence rate metric would be reset to zero. If an item does not converge, the convergence rate for that game would be zero. 

\begin{table}[]
\begin{tabular}{|m{5em}|m{5em}|m{25em}|m{5em}|}
\hline
\textbf{Animal} & \textbf{Coreference resolution} & \textbf{Excerpt extracted}                                                                                                                                                                                                                   & \textbf{Probability} \\ \hline
Cheetah         & No                              & They have been widely depicted in art, literature, advertising, and animation.                                                                                                                                                               & 0.17                 \\ \hline
Cheetah         & No                              & An open area with some cover, such as diffused bushes, is probably ideal for the cheetah because it needs to stalk and pursue its prey over a distance.                                                                                      & 0.10                 \\ \hline
Cheetah         & Yes                             & Generally, the female can not escape on her own; the males themselves leave after they lose interest in her.                                                                                                                                 & 0.41                 \\ \hline
Cheetah         & Yes                             & == Interaction with humans ==\textbackslash{}n\textbackslash{}n\textbackslash{}n=== Taming ===\textbackslash{}n\textbackslash{}n,The cheetah shows little aggression toward humans, and can be tamed easily, as it has been since antiquity. & 0.41                 \\ \hline
Monkey          & No                              & Some are kept as pets, others used as model organisms in laboratories or in space missions.                                                                                                                                                  & 0.24                 \\ \hline
Monkey          & No                              & They are used primarily because of their relative ease of handling, their fast reproductive cycle (compared to apes) and their psychological and  physical similarity to humans.                                                             & 0.16                 \\ \hline
Monkey          & Yes                             & The most common monkey species found in animal research are the grivet, the rhesus macaque, and the crab-eating macaque, which are either wild-caught or purpose-bred.                                                                       & 0.49                 \\ \hline
Monkey          & Yes                             & Some are kept as pets, others used as model organisms in laboratories or in space missions.                                                                                                                                                  & 0.45                 \\ \hline
Elephant        & No                              & In the past, they were used in war; today, they are often controversially put on display in zoos, or exploited for entertainment in circuses.                                                                                                & 0.26                 \\ \hline
Elephant        & No                              & It can be used for delicate tasks, such as wiping an eye and checking an orifice, and is capable of cracking a peanut shell without breaking the seed.                                                                                       & 0.13                 \\ \hline
Elephant        & Yes                             & === Zoos and circuses ===\textbackslash{}n\textbackslash{}nElephants were historically kept for display in the menageries of Ancient Egypt, China, Greece, and Rome.                                                                         & 0.50                 \\ \hline
Elephant        & Yes                             & In the past, they were used in war; today, they are often controversially put on display in zoos, or exploited for entertainment in circuses.                                                                                                & 0.44                 \\ \hline
\end{tabular}
\caption{An example showing the Coreference Resolution Dilemma}
\label{table:coref-resolution-dilemma}
\end{table}

\section{Limitations}
As we defined a new problem in NLP and provided preliminary results for the same, we observed some significant shortcomings in the problem-definition, current state of transformer models, our primary dataset BoolQ, using Wikipedia as our primary corpus, and limitations of word2vec models. While working with general entities, our baseline models failed to understand subtleties as it seemed to require a vast amount of global information to decisively answer ‘no’. Hence, to make the problem tractable, we modified the problem definition to only include animal names as our ‘guess’ words. Furthermore, the transformer models we worked with - DistilBERT and RoBeRTa - showed difficulty in performing comparison and counting tasks. For example, our model would often fail when presented with questions such as ‘Is it smaller than a monkey?’ (comparative type) and ‘Does it have 8 legs?’ (counting type). While a human can visually comprehend such tasks, it becomes difficult to find such sentences in a corpus which can validate the presence of such sentences. Moreover, we believe as a future-scope in the Computer Vision domain, one can include a multimodal pipeline which combines ours with one that performs question-answering by observing an image. 

Another limitation of the transformer model is that the negative results are hard to guess - as mentioned in the BoolQ paper \cite{clark2019boolq}, the subtlety of negation lies in understanding that ‘a positive assertion in the text excludes, or makes unlikely, a positive assertion in the question’. As mentioned in RoBeRTa and DistilBERT sections, another limitation we observed was overfitting during our finetuning on the BoolQ dataset. 

We observed that while BoolQ dataset is modelled to solve a yes/no problem, the subtleties between their and our problem definitions add up significantly. For instance, almost all of our questions start with the word ‘is’, however, more than 50\% of our training data (5234 examples) consists of questions not starting with ‘is’. Furthermore, as mentioned before, many ‘animal’ related questions required prior knowledge of other animals to answer correctly - however, the training corpus was largely devoid of questions from our problem domain. We also observed that both the Spacy and Gensim word2vec models had difficulty understanding the relationship between an animal and its parent class - for example, a ‘tiger’ had a higher correlation with a reptile, than with a carnivore or a mammal. This made it significantly difficult to perform positive sampling, requiring us to utilize UCI’s zoo dataset \cite{Dua:2019} for obtaining the parent-child relationships for positive/negative sampling. Lastly, we would like to stress that in spite of the vast sea of resources in Wikipedia articles, we found many instances in which both the Simple Wikipedia and Full Wikipedia were unable to find a relevant sentence. For example, while tigers can swim well, their Wikipedia article has no such reference of it, which in turn confuses our model which is dependent upon a strong reference to base its answer on.

\section{Applying in practice}
The biggest prerequisites to apply this problem in practice would be to fine-tune the yes/no model on a transformer trained on a larger dataset such as GPT-2 (which has 1.5 billion parameters and was trained on a dataset of 8 million web pages) \cite{gpt}. Another prerequisite would be to build a vast taxonomy to improve the performance of the positive/negative sampling stages of the pipeline. We also propose using a hybrid corpus consisting of answers from Wikipedia and domain-specific knowledge graphs. We observed that knowledge graphs such as DBPedia \cite{auer2007dbpedia} heavily borrowed their content from Wikipedia, making it less effective for this task. Moreover, if the domain problem requires a broader category of entities, we highly suggest creating a custom dataset for your task, instead of overly relying upon BoolQ due to its limitations (as mentioned in the Limitations section). Lastly, if one expects the questions to include more than one pronoun, we encourage building a pronoun resolution model - starting with a baseline model (like the Hobbs’ algorithm) \cite{hobbs1978resolving} and eventually experimenting with Google's GAP dataset \cite{gap}.

\section{Future Work}
It would be helpful if we could detect questions that require real world knowledge to answer. These questions are often in the form of comparisons to other objects/animals such as \textit{Is your animal bigger than a human}? As future work, it would be interesting to identify questions that present a comparison-type query and answer these questions with an \textit{idk} to avoid confusing the model with confident but incorrect answers. The original Akinator tends to guess \textit{a guy who answers randomly} if the model answers \textit{idk}, \textit{probably} or \textit{probably not} too many times. We could maintain a penalty for such answers that increases every time the model outputs an uncertain answer and decreases with every definite answer that the model outputs.

\printbibliography






\appendix

\end{document}